\documentclass[conference]{IEEEtran}
\IEEEoverridecommandlockouts
\usepackage{cite}
\usepackage{amsmath,amssymb,amsfonts}
\usepackage{algorithmic}
\usepackage{graphicx}
\usepackage{textcomp}
\usepackage{subcaption}
\usepackage{xcolor}
\def\BibTeX{{\rm B\kern-.05em{\sc i\kern-.025em b}\kern-.08em
    T\kern-.1667em\lower.7ex\hbox{E}\kern-.125emX}}
\begin{document}

\title{Assessing a Single Student's Concentration on Learning Platforms: A Machine Learning-Enhanced EEG-Based Framework\\
\thanks{----------------------------------
\newline
\hspace*{0.4cm}\textsuperscript{*}Corresponding Author}
}

\author{
\begin{minipage}{0.35\textwidth}
    \centering
    1\textsuperscript{st} Zewen Zhuo \\
    \textit{Faculty of Science and Technology} \\
    \textit{University Paris-Est Créteil}\\
    Créteil, France \\
    zewen.zhuo@etu.u-pec.fr
\end{minipage}
\begin{minipage}{0.35\textwidth}
    \centering
    2\textsuperscript{nd} Mohamad Najafi \\
    \textit{Faculty of Science and Technology} \\
    \textit{University Paris-Est Créteil}\\
    Créteil, France \\
    mohamad.najafi@etu.u-pec.fr
\end{minipage}
\begin{minipage}{0.3\textwidth}
    \centering
    3\textsuperscript{rd} Hazem Zein \\
    \textit{LISSI Laboratory} \\
    \textit{University Paris-Est Créteil}\\
    Vitry-sur-Seine, France \\
    hazem.zein@u-pec.fr
\end{minipage}

\\
\\

\begin{minipage}{0.3\textwidth}
    \centering
    4\textsuperscript{th *} Amine Nait-Ali \\
    \textit{LISSI Laboratory} \\
    \textit{University Paris-Est Créteil}\\
    Vitry-sur-Seine, France \\
    naitali@u-pec.fr
\end{minipage}
}

\maketitle

\begin{abstract}
This study introduces a specialized pipeline designed to classify the concentration state of an individual student during online learning sessions by training a custom-tailored machine learning model. Detailed protocols for acquiring and preprocessing EEG data are outlined, along with the extraction of fifty statistical features from five EEG signal bands: alpha, beta, theta, delta, and gamma. Following feature extraction, a thorough feature selection process was conducted to optimize the data inputs for a personalized analysis. The study also explores the benefits of hyperparameter fine-tuning to enhance the classification accuracy of the student’s concentration state. EEG signals were captured from the student using a Muse headband (Gen 2), equipped with five electrodes (TP9, AF7, AF8, TP10, and a reference electrode NZ), during engagement with educational content on computer-based e-learning platforms. Employing a random forest model customized to the student’s data, we achieved remarkable classification performance, with test accuracies of 97.6\% in the computer-based learning setting and
98\% in the virtual reality setting. These results underscore the effectiveness of our approach in delivering personalized insights into student concentration during online educational activities.

\end{abstract}


\section{Introduction}
In the past few years, e-learning has gained more attention due to the global COVID-19 pandemic, during which online education became the primary method of imparting knowledge. Nowadays, hybrid learning remains a prevalent modality in many universities and institutions because of its benefits such as low cost, unlimited access, and the availability of comprehensive materials. Notably, with the advancement of virtual reality (VR) systems, an increasing number of users have switched their learning format from traditional e-learning devices, like personal computers and tablets, to VR for a more immersive experience. Understanding learners’ mental states during online study sessions can be complicated but is necessary for lecturers to adjust the curriculum and assist students with difficulties\cite{b1}. One effective method for assessing human concentration is electroencephalography (EEG), a noninvasive technique that collects informative brain data via electrodes placed on the scalp. Typically, EEG signals are segmented into five bands based on their frequency range: delta ($\delta$) for frequencies less than 4 Hz, theta ($\theta$) for 4-7 Hz, alpha ($\alpha$) for 8-12 Hz, beta ($\beta$) for 13-30 Hz, and gamma ($\gamma$) for frequencies above 30 Hz\cite{b2}. Although understanding attentional states requires expertise in various fields, including sensor technology and neurophysiology, the rapid advancement of machine learning (ML) technology has facilitated scientists in analyzing these complex signals.

The deployment of wearable EEG devices has gained significant popularity among research groups worldwide for classifying cognitive states and other purposes. Classification between concentration and meditation using a random forest classifier \cite{b3} explored different statistical features and manifested an accuracy of 75\%. It demonstrates the robustness of utilizing wearable EEG devices and ML models for mental state classification. Mental state recognition in real-time in the e-learning environment \cite{b4} depicted a promising perspective of maximizing users' learning performance in the adaptive format using advanced ML techniques. Multi-class classification among relaxing, neutral, and concentrating through a combination of features and classification models like Bayesian networks, support vector machines, and random
forest \cite{b5} leveraged the overall classification accuracy to 87\% and unleashed the potential of mental state classification with higher accuracy by sophisticated feature selection. 

A major challenge of EEG data classification is extracting meaningful features and mapping them to multiple categories. The current research mainly focuses on the binary classification of EEG signals, which is insufficient for understanding online learning performance. Furthermore, the lack of comparison of model accuracy with different numbers of selected features impedes efforts to maximize model performance. Furthermore, the investigation on VR-based e-learning is still in its early stages, requiring more study to integrate VR in online learning concentration estimation. To address these issues, we propose a pipeline to develop ML models for individual user-based concentration state classification in both VR and non-VR learning environments. It provides a thorough approach starting from data collection to feature selection and model fine-tuning, applying the most effective generic method to analyze the attentional states in different learning formats. This method not only enhances the performance of ML models in multi-class mental state classification but also opens new avenues for comparing different e-learning formats.

 This paper is structured as follows: the experimental setup, methodology, and specifications regarding the deployed devices are summarized in Section \ref{2}. Section \ref{3} manifests the detailed results as well as comparison, performance evaluation, and some further discussions. A final conclusion and other future perspectives are provided in Section \ref{4}.

\begin{figure}[ht]
    \centering

    \begin{minipage}{0.48\textwidth}
        \centering
        \includegraphics[width=\linewidth]{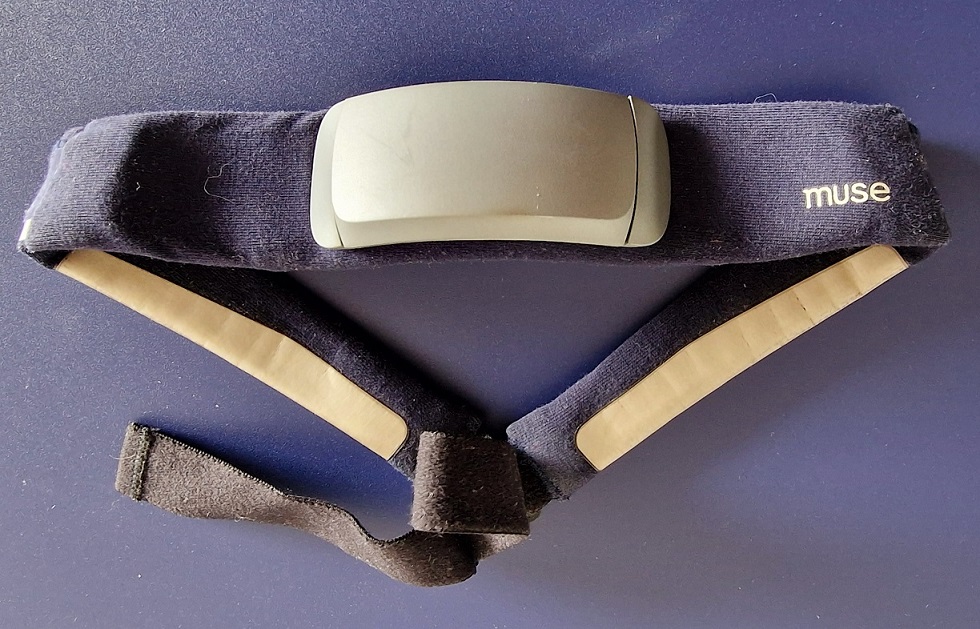}
    \end{minipage}%
    \hfill
    \begin{minipage}{0.48\textwidth}
        \centering      \includegraphics[height=5.5cm,width=\linewidth]{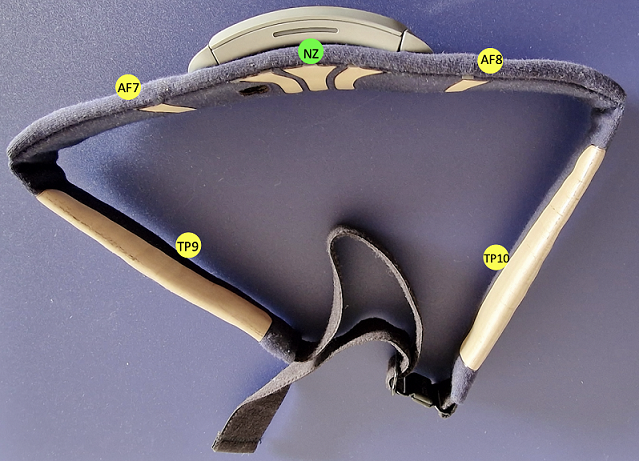}
    \end{minipage}

    \caption{\textbf{\textit{Muse S} EEG Headband}: the green and yellow labels correspond to the international 10-20 EEG electrode placement standard}
    \label{fig:complete}
\end{figure}

\section{Materials and Methodology}\label{2}
\subsection{EEG Data Acquisition}
\begin{figure}[ht]
\centering
\includegraphics[width=\linewidth]{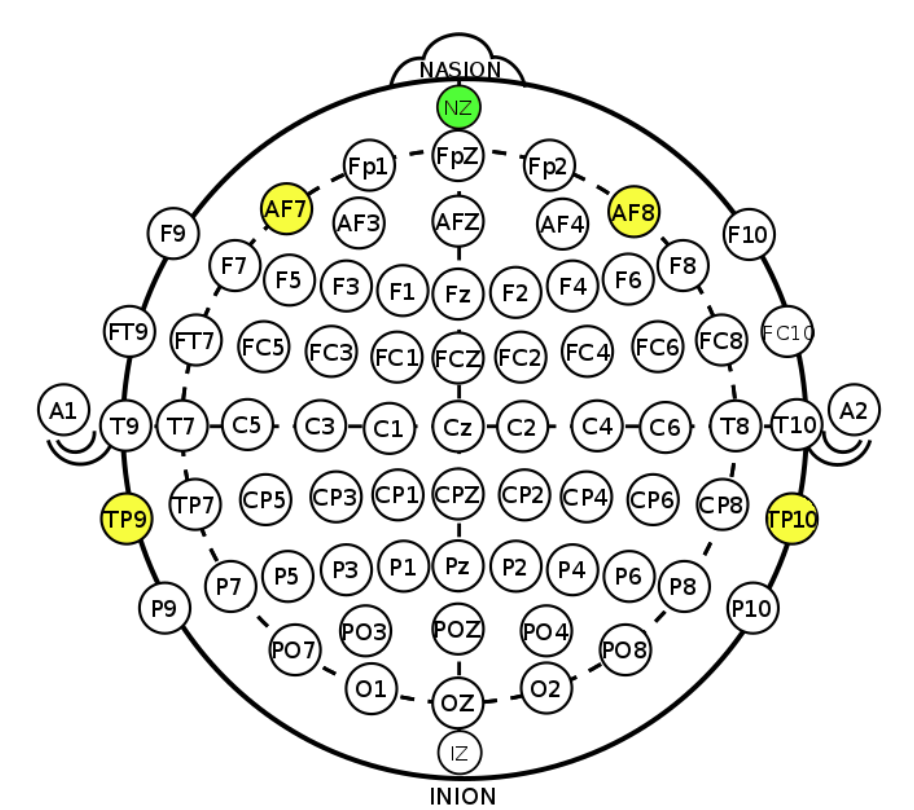}
\caption{\textbf{The International 10-20 EEG Electrode Placement Standard }\cite{b5}: the yellow labels are electrodes of \textit{Muse} headband and green labeled sensor is deployed as a reference point for calibration}
\label{fig2}
\end{figure}

The study aims to provide guidelines for achieving ML-based concentration classification in two different e-learning formats utilizing EEG data. Traditional clinical-level EEG setups are expensive and hard to operate without expertise \cite{b6,b7,b8}. In this research, we utilized the \textit{Muse S (Gen 2)}, as shown in Fig. \ref{fig:complete}, which has been employed in diverse research circumstances\cite{b9,b10,b11}. This portable and lightweight headband is a commercial EEG sensing equipment, embedded with five sensors. Each sensor is positioned in the specific region as illustrated in Fig. \ref{fig2}, including TP9, AF7, AF8, and TP10, to record brain activities, along with a reference point (NZ) to calibrate the brain signals. In most clinical settings, EEG signals are recorded as raw data from each electrode. However, these signals are typically intricate and challenging to interpret without the assistance of specialists. One effective way to simplify the analysis is to decompose the raw EEG signals into five distinct frequency bands through the application of band-pass filters\cite{b12}. Besides that, many studies stated that the power line interference can significantly undermine EEG signal analysis\cite{b13,b14}. The common practice is to apply a notch filter (50 or 60 Hz, depending on the geographic region) to phase out undesired power noise, thereby mitigating signal contamination. The availability of \textit{Mind Monitor} on various mobile devices and its capacity to apply notch and band-pass filters made it ideal for data acquisition. Furthermore, it can record the processed signals. During data collection, a sampling frequency of 256 Hz was selected to ensure the capture of meaningful features for the classification task.

Aiming to provide generic model development pipelines for both conventional e-learning (e.g. on personal computers) and VR-based e-learning, we utilized the \textit{Meta Quest 2} headset to offer an engaging and intellectually stimulating learning experience for the participant.

\subsection{Protocols and Methodology}
To explore the viability of user-based model design and its pipeline, EEG data were acquired from a single participant. The candidate, a master's student in their 20s, voluntarily participated in this study and provided informed consent. While conducting the experiment, different education-oriented videos, with a duration of approximately 5 minutes for each, were selected in both learning modalities. These videos were related to the candidate's educational background and the participant watched these videos with time intervals between each session. While watching, \textit{Muse S} headband was placed in the correct position to ensure that EEG signals were recorded simultaneously. After each video, the participant assessed their own concentration levels, classifying them into three categories: \textit{fully concentrated}, \textit{moderately concentrated}, and \textit{not concentrated}. In total, 12 different videos were used, with 6 videos on VR and 6 videos on computers (hereafter referred to as "VR sessions" and "non-VR sessions" respectively). The sampling frequency was set to 256 Hz for all sessions with a 50 Hz notch filter applied.

\begin{figure}[h]
\centering
\includegraphics[height=6cm,width=\linewidth]{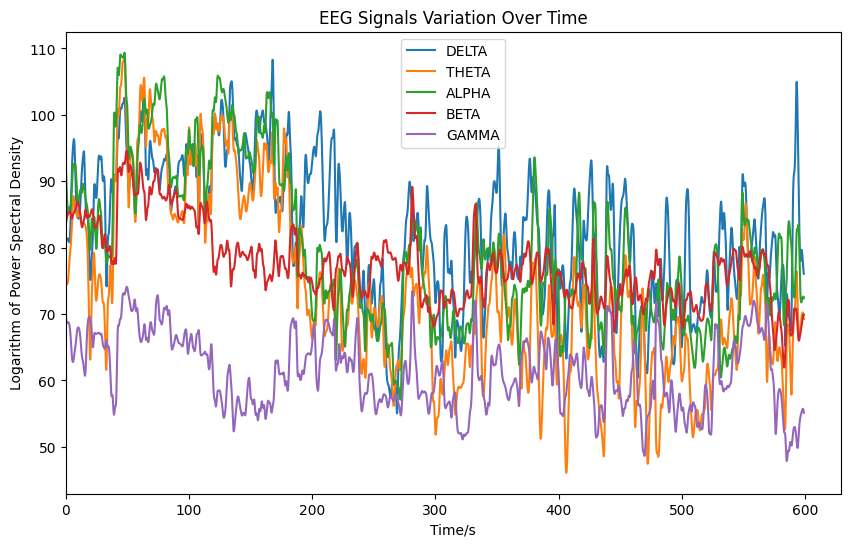}
\caption{\textbf{Example of EEG  Variations over Time}: the x-axis indicates time (in seconds), and the y-axis represents the logarithm of power spectral density across different frequency bands (delta, theta, alpha, beta, gamma)}
\label{fig3}
\end{figure}

Upon completing data collection, the raw data comprised 20 key values: the logarithm of power spectral density for 5 signal bands, gathered from 4 nodes. To better represent the variation and simplify the task, mean value of each signal band was calculated with Eq. (\ref{eq:mean_formula}), where \textit{X} represents frequency band, and rescaled. An example of EEG signals recorded in this study is presented in Fig. \ref{fig3}.

\begin{equation}
\label{eq:mean_formula}
\textit{value} = \frac{{X_{\textit{AF7}} + X_{\textit{AF8}} + X_{\textit{TP9}} + X_{\textit{TP10}}}}{\textit{4}}
\end{equation}
In addition, bad connection points were handled by forward filling, taking the value from the previous collected point. 

After data underwent pre-processing, the following step was to extract statistical features. In this study, mean value, squared value, variance, standard deviation, skewness, kurtosis, root mean square, entropy, activity, and mobility were extracted for each frequency band of EEG signals. Subsequently, features acted as inputs to 5 ML models, namely Support Vector Machine (SVM), Dense Neural Network (DNN), Random Forest (RF), Adaptive Boosting (AdaBoost), and Extreme Gradient Boosting (XgBoost), together with top-k feature selection and model hyper-parameter fine-tuning. In our study, due to the nature of the data and the lack of adequacy, Convolutional Neural Networks (CNNs) were not considered.

\subsection{Feature Extraction}
Unlike CNNs, which are capable of extracting features from images through a deep network with tremendous layers, extracting informative features is crucial for effective ML models in the traditional paradigm. In this investigation, we took the strategy of extracting as many features as possible, followed by a feature selection approach, to determine the best number of features for the individual model. All models tested fall into the bracket of supervised learning category and the details of feature-label pairs are summarized in Table \ref{tab1}. Overall, fifty features were distilled from the pre-processed signals with their respective labels. Through successive analysis, a sliding window size of 10 seconds was implemented for feature extraction in both non-VR and VR sessions, resulting in 3580 sample points in the former modality and 3582 points in the later modality (Table \ref{table:sample_distribution}), preserving the data balance.

\begin{table}[htbp]
\caption{Extracted Statistic Features and User Feedback Labels}
  \begin{center}
\begin{tabular}{|p{4cm}|p{3cm}|}
  \hline
  \textbf{Features$^{\mathrm{a}}$} & \textbf{Labels} \\ \hline
  \multicolumn{1}{|p{4cm}|} {\textit{squared value\textsubscript{i}}\hspace*{0.8cm}\textit{mean value\textsubscript{i}}} & \textit{Fully Concentrated} \\
  \multicolumn{1}{|p{4cm}|} {\textit{variance \textsubscript{i}}\hspace*{0.6cm}\textit{standard deviation \textsubscript{i}}} & \\ 
  \multicolumn{1}{|p{4cm}|} {\textit{skewness\textsubscript{i}}\hspace*{1.8cm}\textit{kurtosis\textsubscript{i}}} & \textit{Moderately Concentrated} \\ 
  \multicolumn{1}{|p{4cm}|} {\textit{root mean square\textsubscript{i}}\hspace*{0.8cm}\textit{entropy\textsubscript{i}}} & \\ 
  \multicolumn{1}{|p{4cm}|} {\textit{activity\textsubscript{i}}\hspace*{1.9cm}\textit{mobility\textsubscript{i}}} & \textit{Not Concentrated} \\ 
 \hline
    \multicolumn{2}{p{7cm}}{$^{\mathrm{a}}$ Where \textit{i} represents 5 different EEG frequency bands}
  \end{tabular}
     \end{center}
\label{tab1}
\end{table}

\begin{table}[h]
\centering
\caption{Distribution of Sample Number in Training, Validation, and Test Subsets in Non-VR and VR Sessions}
\begin{tabular}{|p{1cm}|c|c|c|c|}
\hline
 & \textbf{Sample Number} & \textbf{Training} & \textbf{Validation} & \textbf{Test} \\ \hline
\textbf{Non-VR} & 3580 & 2146 (60\%) & 714 (20\%) & 720 (20\%) \\ \hline
\textbf{VR} & 3582 & 2148 (60\%) & 714 (20\%) & 720 (20\%) \\ \hline
\end{tabular}
\label{table:sample_distribution}
\end{table}

\section{Results and Discussion}\label{3}

\subsection{Model Training and Feature Selection}

When the expected features were ready, the data were subsequently split into three subsets: a training subset (60\%), a validation subset (20\%), and a test subset (20\%), as outlined in detail in Table \ref{table:sample_distribution}. During the splitting process, data from independent recordings were not mixed together due to their sequential nature, ensuring that each recording was sampled equally. The split data were then merged and shuffled to maintain randomness. By default, all existing features were used as input for five models, setting the benchmark accuracy and laying the foundation for later performance enhancements. The default validation accuracy is reported in  Tabel \ref{tab2} and \ref{tab3}.

\begin{figure}[h]
  \centering
  \begin{subfigure}[b]{0.45\textwidth}
    \includegraphics[width=\textwidth]{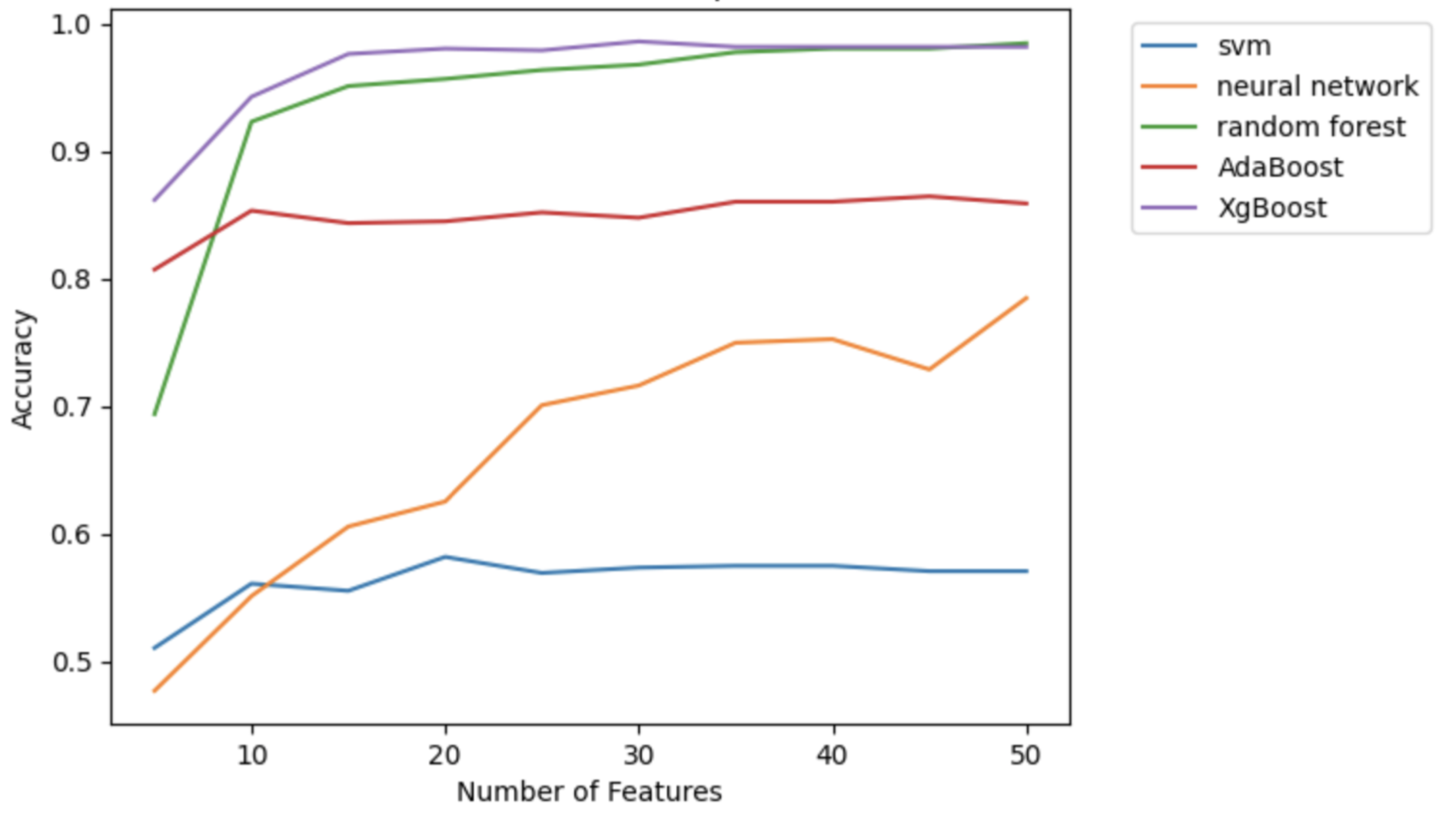}
    \caption{Non-VR Validation Accuracy over Feature Number}
    \label{fig:sub1} 
  \end{subfigure}
  \hfill 
  \begin{subfigure}[b]{0.435\textwidth}
    \includegraphics[width=\textwidth]{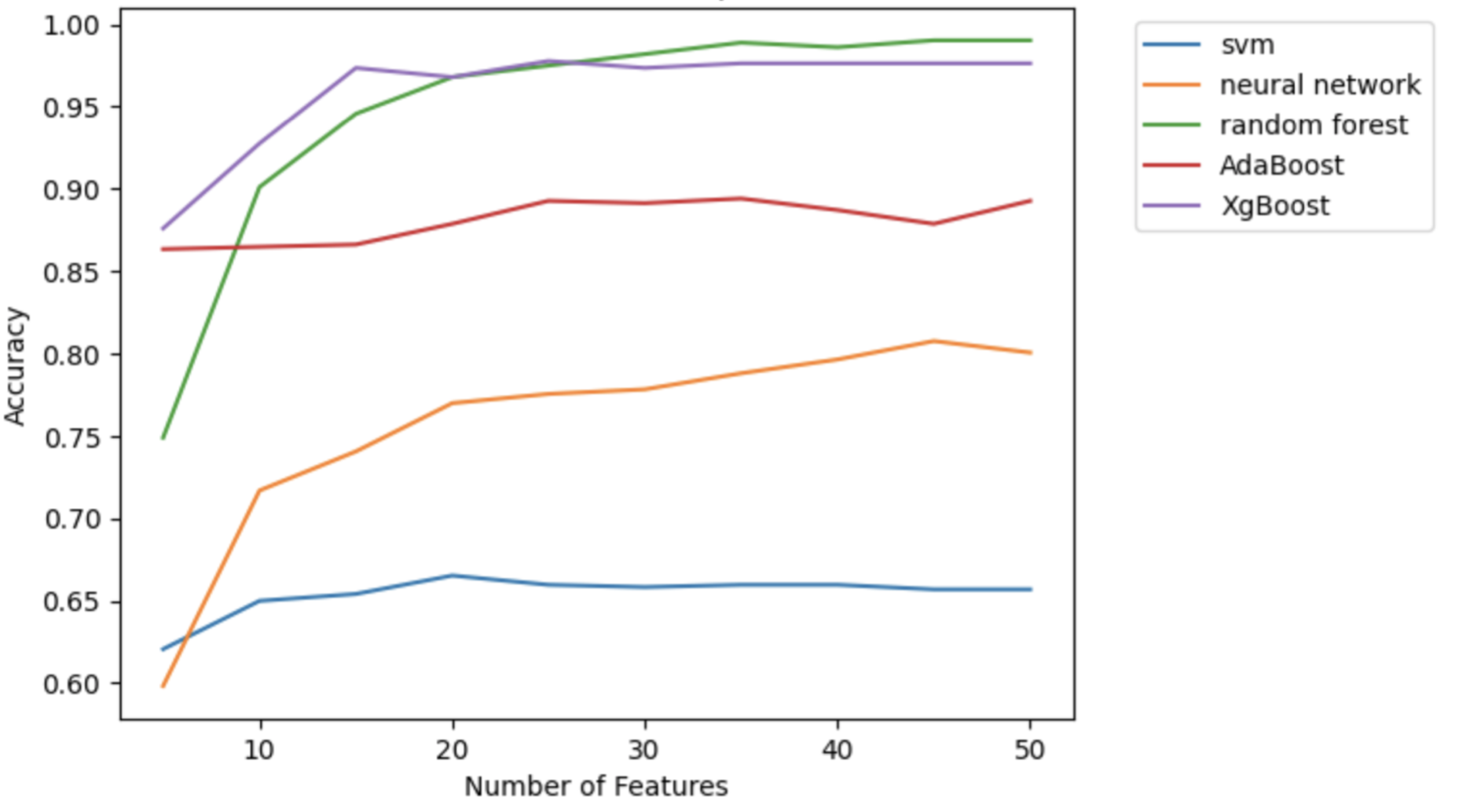}
    \caption{VR Validation Accuracy over Feature Number}
    \label{fig:sub2}
  \end{subfigure}
  \caption{\textbf{Validation Accuracy Variation}: the x-axis represents the number of selected features, with an interval of 5, and y-axis stands for validation accuracy}
  \label{fig6} 
\end{figure}

Nevertheless, not all features are informative and valuable for these models, indicating the necessity of feature selection. The top-k feature selection strategy based on feature importance was employed, with 5 features as the interval. The variation in validation accuracy over the number of features for non-VR and VR sessions is depicted in Fig. \ref{fig:sub1} and \ref{fig:sub2}.

    In Fig. \ref{fig:sub1}, the validation accuracy for non-VR data is plotted against the number of selected features. The SVM model shows the least sensitivity (accuracy improvement $< $ 1\%) to the number of features, with its accuracy remaining around 55\%. In contrast, the RF and XgBoost models show significant improvements, reaching validation accuracies above 98\% with an optimal number of features. The DNN and AdaBoost models also benefit from feature selection, achieving higher accuracies as the number of features increases. In Fig. \ref{fig:sub2}, it is intuitive to find that, similar to the non-VR data, the SVM model remains relatively unaffected by the number of features, with an accuracy of around 65\%. The RF and XgBoost models once again manifest superior performance, achieving validation accuracies near or at 99\% with an optimal feature set. The DNN and AdaBoost models also exhibit improved performance with an increasing number of features.
    \begin{table}[h]
\caption{Models Performance on non-VR Validation Subset}
\begin{center}
\begin{tabular}{|p{1.9cm}|p{1cm}|p{1.85cm}|p{1.3cm}|p{0.8cm}|}
\hline
\textbf{Models} & \textbf{All Features} & \textbf{Number of Selected Features} & \textbf{After Selection} & \textbf{Fine-Tuned} \\ \hline
SVM              & 57.1\%              &20              &58.2\%            & 94.5\%            \\
Neural Network  & 78.5\%    &50 &78.5\%        & 96.4\%            \\
\textbf{Random Forest}    & \textbf{98.3}\%  &\textbf{50} &\textbf{98.3}\%          & \textbf{98.3}\%            \\
AdaBoost         & 84.2\%   &45 &86.2\%         & 97.9\%            \\
XgBoost          & 97.2\%    &30 &98.6\%        & 98.6\%            \\
\hline
\end{tabular}
\end{center}
\label{tab2}
\end{table}

\begin{table}[h]
\caption{Models Performance on VR Validation Subset}
\begin{center}
\begin{tabular}{|p{1.9cm}|p{1cm}|p{1.85cm}|p{1.3cm}|p{0.8cm}|}
\hline
\textbf{Models} & \textbf{All Features} & \textbf{Number of Selected Features} & \textbf{After Selection} & \textbf{Fine-Tuned} \\ \hline
SVM              & 65.7\%              &20              &66.5\%            & 91.4\%            \\
Neural Network  & 80.1\%    &45 &80.8\%        & 91.9\%            \\
\textbf{Random Forest}    & \textbf{98.9}\%  &\textbf{45} &\textbf{99.0}\%          & \textbf{99.0}\%            \\
AdaBoost         & 88.0\%   &35 &89.4\%         & 95.0\%            \\
XgBoost          & 97.8\%    &25 &97.8\%        & 97.8\%            \\
\hline
\end{tabular}
\end{center}
\label{tab3}
\end{table}

    Tables III and IV provide a detailed comparison of model performances on non-VR and VR validation subsets, respectively. They showcase the accuracy of each model when using all features, after feature selection, and after hyper-parameter fine-tuning, with the optimal number of features reported. The XgBoost model achieves the highest validation accuracy after feature selection and fine-tuning for non-VR data (Table \ref{tab2}) at 98.6\%, 0.3\% better than RF, while for VR data (Table \ref{tab3}), the RF model reaches an accuracy of 99.0\%. Notably, the RF model demonstrates consistently high performance across both settings, despite being marginally smaller than the XgBoost in non-VR sessions. Therefore, it has been chosen as the model to be evaluated on the test subset.
    \begin{figure}[h]
  \centering
  \begin{subfigure}[b]{0.45\textwidth}
    \includegraphics[width=\textwidth]{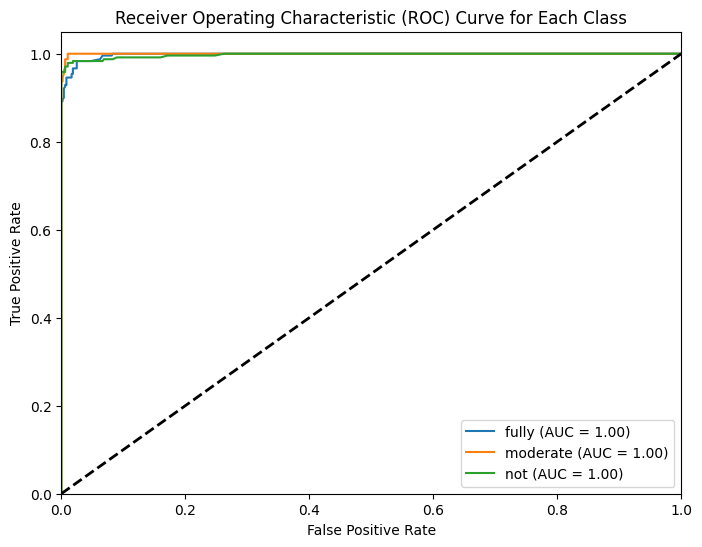}
    \caption{ROC Curve of RF on non-VR Test Subset}
    \label{fig:sub_a} 
  \end{subfigure}
  \hfill 
  \begin{subfigure}[b]{0.435\textwidth}
    \includegraphics[width=\textwidth]{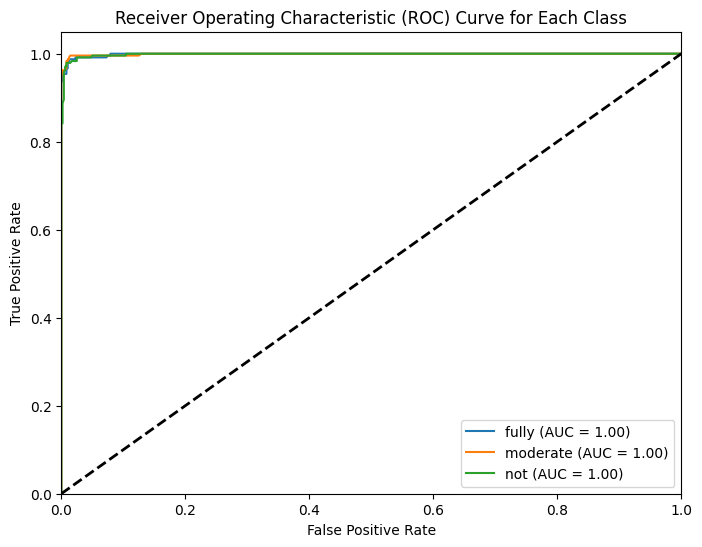}
    \caption{ROC Curve of RF on VR Test Subset}
    \label{fig:sub_b} 
  \end{subfigure}
  \caption{\textbf{Generalizability Representation}: these ROC curves illustrate the performance of the RF model on the (a) non-VR and (b) VR test subset, achieving an AUC of 1.00 for each class}
  \label{fig7} 
\end{figure}

\subsection{Evaluation}
    Following validation, the RF model was tested on unseen data, the test subset, to evaluate its robustness and stability. The first evaluation technique utilized was receiver operating characteristic (ROC) curves for each class, which represent the generalizability of the model on unseen data. The final results, as shown in Fig. \ref{fig7}, demonstrate the model’s impressive generalization on unseen data. The ROC curves (Fig. \ref{fig:sub_a} and \ref{fig:sub_b}) indicate that the RF model achieved an area under the curve (AUC) of 1.00 for all classes in both non-VR and VR sessions, suggesting outstanding discrimination between the classes. This high AUC value affirms the model’s excellent performance and robustness.

\begin{figure}
  \centering
  \begin{subfigure}[b]{0.45\textwidth}
    \includegraphics[width=\textwidth]{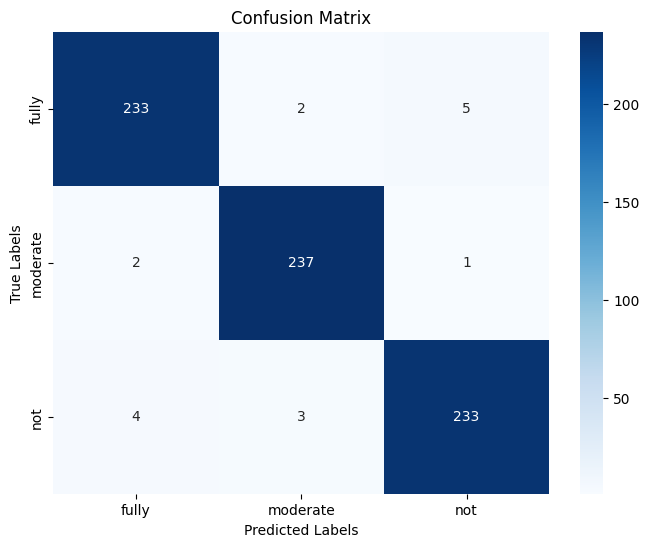}
    \caption{Confusion Matrix of RF on non-VR Test Subset}
    \label{fig:sub_aa} 
  \end{subfigure}
  \hfill 
  \begin{subfigure}[b]{0.435\textwidth}
    \includegraphics[width=\textwidth]{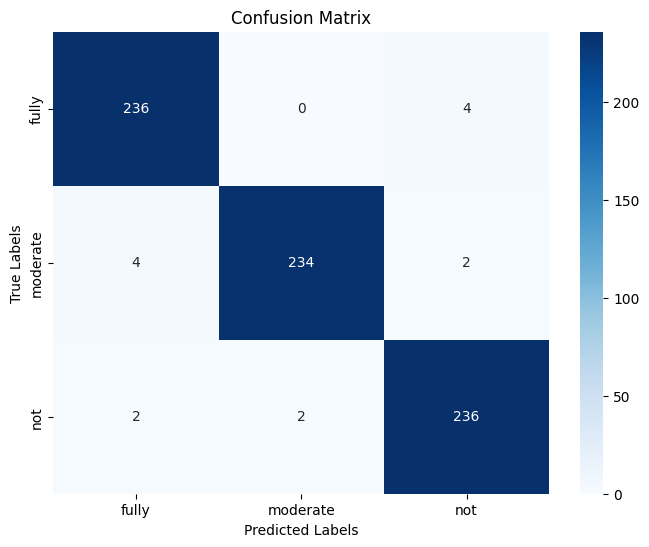}
    \caption{Confusion Matrix of RF on VR Test Subset}
    \label{fig:sub_bb} 
  \end{subfigure}
  \caption{\textbf{Confusion Matrices}: these confusion matrices showcase the classification performance of the RF on the nonVR and VR test subset in (a) and (b) respectively.}
  \label{fig8} 
\end{figure}

    Beyond that, the model was evaluated using a confusion matrix on the test subset of data. The outcomes of the confusion matrices (Fig. \ref{fig:sub_aa} and \ref{fig:sub_bb}) imply that the chosen model can classify the attentional states of sample points with an accuracy higher than 90\%, specifically 97.6\% for non-VR and 98\% for VR, which is remarkable compared to relevant studies \cite{b2,b3,b4,b11,b12}, further solidifying the model’s efficacy.

\subsection{Discussion}
This study demonstrates a clear pipeline for developing user-based attentional state classification models in educational settings. The proposed model achieved high accuracy (97.6\% for non-VR and 98\% for VR) in differentiating between \textit{fully concentrated}, \textit{moderately concentrated}, and \textit{not concentrated} states. This high level of accuracy underscores the model's potential utility in real-world educational applications, providing a valuable tool for monitoring and enhancing student engagement.

However, a key limitation of the study lies in the lack of participant diversity. The relatively small and homogeneous sample suppresses the generalizability of the findings and the robustness of the pipelines. This constraint undermines the comparison of features and models selected for different users, potentially affecting the final results.

The absence of a diverse candidate pool underlines the significance of data diversity in robust model development. To address this limitation, future work can focus on expanding the dataset by recruiting participants from various demographics, including different majors, age groups, genders, and ethnicities. A more diverse dataset would provide a comprehensive understanding of how different factors influence feature and model selection.

Additionally, future research should explore the implementation of the model development pipeline across a broader range of educational tasks and environments, more cognitive states, and with finer feature selection intervals. This would help in assessing the adaptability and effectiveness of the model in different contexts, ensuring its applicability in diverse real-world scenarios.

\section{Conclusion}\label{4}
In conclusion, this study presents a pipeline for classifying individual student concentration states into three classes during online learning using ML techniques applied to EEG data. By pre-processing, statistical features extraction, performing feature selection, and fine-tuning hyper-parameters, we achieved highly accurate classification of cognitive states. Our findings indicate that the user-based RF model, trained on data collected from the \textit{Muse} headband across both computer-based and VR learning environments, yielded exceptional classification performance, with test accuracies surpassing 97\%. These results suggest the potential of EEG-based methods to analyze and understand personalized learning experiences. They offer a detailed statistical analysis of different engagement levels across various modalities, enhancing the personalization of learning. Future research may explore the integration of such techniques into educational platforms to optimize instructional design and support individualized learning pathways.

\end{document}